# Cost-Penalized Fitness in FMA-Orchestrated Mixture of Experts:
# Experimental Evidence for Molecular Memory in Domain Adaptation

Martin Jaraíz

*Department of Electricity and Electronics, University of Valladolid, Spain*

April 2026

**Abstract**

We present experimental results from seven controlled runs of nanoFMT, a Free-Market Algorithm (FMA) orchestrated transformer with dynamic Mixture-of-Experts (MoE) management. The experiments address a fundamental question for advanced LLM development: how should an MoE system manage its expert pool when operating at full capacity under changing data distributions? We demonstrate that cost-penalized fitness metrics, combined with a linear grace period for newborn experts, produce a system that accumulates domain expertise through diversification rather than replacement. The central result is a round-trip domain shift experiment showing **9–11× faster recovery** when returning to a previously learned domain, with **zero expert births or replacements** required. This "molecular memory" effect — where dormant experts survive and reactivate when their domain returns — has no analogue in current MoE management approaches. A preliminary cost analysis estimates annual savings of **$39.1M** and **27.1 GWh** energy reduction for an OpenAI-scale provider under a moderate scenario.

## 1. Introduction

The Free-Market Algorithm (FMA) [1,10] is a universal optimization framework comprising 18 interacting mechanisms — birth, death, fitness evaluation, mutation, migration, and others — that govern populations of competing agents through market-like dynamics. Originally developed for agent-based economic modeling [2], the FMA has since been applied across domains ranging from macroeconomic forecasting [3] to prebiotic chemistry, where its mechanisms find natural analogues in chemical kinetics (reaction intermediates playing the role of economic agents). In this work we apply the FMA to orchestrate a dynamic Mixture-of-Experts (MoE) transformer, yielding a system we call *nanoFMT*.

### 1.1 The Saturated Regime Problem

Production MoE systems — GPT-4 (OpenAI, 2023), DeepSeek-V3 (DeepSeek, 2024), Mixtral (Mistral AI, 2024) — operate at full expert capacity from the start of training. Unlike idealized experimental settings where expert pools grow gradually from empty slots, real systems begin with all expert slots populated and must adapt through *replacement* rather than addition.

Our initial saturated-regime experiments confirmed this challenge: with all 8 expert slots per layer filled from step 0, the FMA's forced replacement mechanism never fired. All experts achieved similar fitness values (Q × D ≈ same), producing near-zero fitness variance. The "worst" expert never crossed the replacement threshold of mean − 1σ. If all experts are equally mediocre, the system has no signal for which ones to replace when the data distribution changes.

## 1.2 The Cost Dimension

The resolution came from an insight rooted in economics: fitness should not measure quality alone. In a competitive market, a successful firm is not simply one with high revenue but one with high revenue *relative to its costs*. Chemistry provides a neat homology: an essential reaction intermediate is not merely one that produces good yield, but one that does so efficiently, consuming fewer resources per unit of product.

Applied to MoE experts: **Fitness = (Quality × Demand) / Compute Cost**. An expert that produces decent routing scores but consumes disproportionate FLOPs should be vulnerable to replacement by a leaner competitor. This cost dimension creates the fitness disparity that the saturated regime suppresses.

## 1.3 Three Notions of Essentiality

Drawing on the chemical homology to the FMA, we identified three complementary notions of what makes an expert "essential":

| Type | Definition | Chemical Analogue |
| --- | --- | --- |
| **Demand-essential** | Routed many tokens (high D) | Molecule present in many pathways |
| **Quality-essential** | Low loss on its tokens (high Q) | Molecule on the shortest pathway |
| **Exclusive-essential** | Sole good choice for certain tokens (high E) | Sole intermediate for a critical product |

These map to the three fitness modes tested: A (Q×D), B (Q×D/FLOPs), and C (Q×D×E/FLOPs).

## 1.4 Related Work

Mixture-of-Experts architectures have evolved from the original top-$k$ gating of Shazeer et al. [7], through Switch Transformers' single-expert routing [6], GShard's capacity-constrained load balancing [8], and Expert Choice routing where experts select tokens rather than vice versa [9]. DeepSeek-V3 [4] introduced auxiliary-loss-free load balancing with 256 experts and 8 active per token. Mixtral [5] demonstrated that moderate-scale MoE (8×7B) could match dense models at a fraction of inference cost.

All these approaches treat the expert pool as *static*: experts are initialized at the start of training and remain fixed throughout. Management reduces to routing optimization — deciding which tokens go to which experts — plus optional post-hoc techniques such as expert pruning (permanently removing low-utility experts), expert merging (combining redundant experts into one), or quantization. None of these methods address the *lifecycle* of experts: when should an expert be born, when should it die, and how should the system manage the transition? More critically, none preserve dormant expertise for later reactivation.

The FMA-orchestrated approach differs fundamentally in that it treats expert management as a *market optimization problem* with birth, death, fitness evaluation, and replacement governed by economic competition. This enables behaviors — molecular memory, infant protection, cost-driven replacement — that have no counterpart in current MoE practice.

## 2. Experimental Design

### 2.1 The nanoFMT Framework

All experiments used nanoFMT, a small-scale FMA-orchestrated transformer [1,2,10] with dynamic expert management, designed for rapid prototyping of MoE governance mechanisms. The FMA's 18-mechanism framework orchestrates expert birth, death, fitness evaluation, and replacement through market-like competition.

### 2.2 Configuration

| Parameter | Value |
|---|---|
| Experts per layer | 8 (saturated from step 0) |
| Layers | 2 |
| Warmup steps | 500 |
| Market evaluation interval | Every 10 steps |
| Domain shift 1 | Step 1500: data_char → data_code |
| Domain shift 2 (c-variants) | Step 3000: data_code → data_char |
| Grace period (b/c-variants) | 50 market steps, linear ramp |
| Replacement threshold | Worst below mean − 1σ |

### 2.3 Fitness Formulas

$$\text{Mode A:} \quad F_i = Q_i \times D_i$$
$$\text{Mode B:} \quad F_i = (Q_i \times D_i) / (1 + \text{FLOPs}_i / \overline{\text{FLOPs}})$$
$$\text{Mode C:} \quad F_i = (Q_i \times D_i \times E_i) / (1 + \text{FLOPs}_i / \overline{\text{FLOPs}})$$
$$\text{Grace:} \quad F^{\text{eff}}_i = F_i \times \min(1, \text{age}_i / \text{grace\_steps})$$

where $Q_i$ is the expert's quality (inverse mean per-token loss), $D_i$ is the fraction of tokens routed to expert *i*, $E_i$ is the exclusivity score (fraction of tokens for which the expert is the top choice by a margin > 0.1), and $FLOPs_i = 2 \times params_i \times tokens\_processed_i$ is a compute cost proxy.

## 2.4 Seven Experimental Variants

| Run | Fitness | Grace | Domain Shifts | Purpose |
|---|---|---|---|---|
| **A2** | Q×D | No | 1500→code | Baseline |
| **B2** | Q×D / FLOPs | No | 1500→code | Cost-penalized |
| **C2** | Q×D×E / FLOPs | No | 1500→code | Exclusivity + cost |
| **B2b** | Q×D / FLOPs | Yes | 1500→code | Cost + infant protection |
| **C2b** | Q×D×E / FLOPs | Yes | 1500→code | Exclusive + infant protection |
| **B2c** | Q×D / FLOPs | Yes | 1500→code→3000→char | Round-trip: cost |
| **C2c** | Q×D×E / FLOPs | Yes | 1500→code→3000→char | Round-trip: exclusive |

# 3. Results

## 3.1 Phase 1: Cost Penalization Creates Fitness Disparity

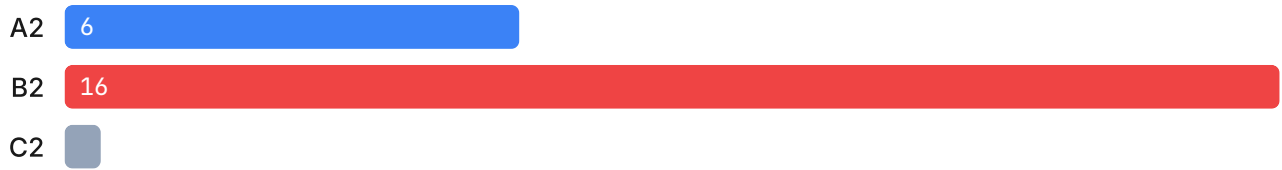

**Figure 1.** Forced replacement counts for Phase 1 (no grace period). Mode B produces a runaway replacement loop (16 events, steps 570–640). Mode C is entirely frozen (0 events). Mode A provides the baseline (6 events).

Cost penalization (Mode B) created sharp fitness differences, but newly spawned experts were killed before they could learn — expert IDs cycled from 13 through 21 in rapid succession. This is the MoE analogue of a market with no infant industry protection. Mode C's exclusivity term compressed all fitness values to F ≈ 0.003–0.005, preventing any expert from crossing the replacement threshold.

## 3.2 Phase 2: Grace Period Balances the Market

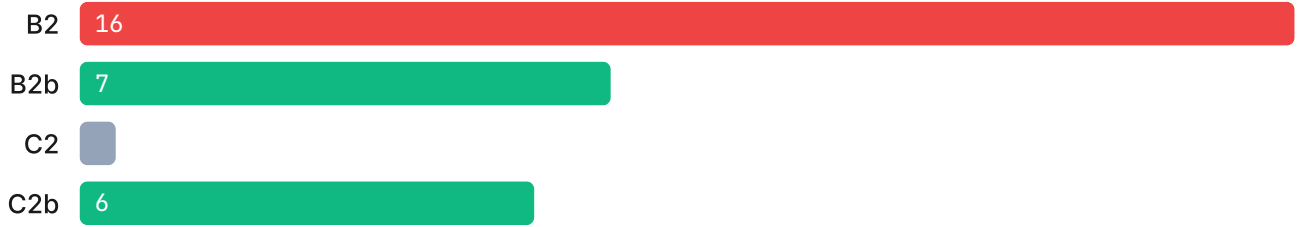

**Figure 2.** Effect of the linear grace period (50 market steps). B2b: runaway loop tamed, 16→7 (−56%). C2b: frozen mode unlocked, 0→6. Post-domain-shift (step 1500+): all checks → KEPT in both runs.

The grace period had a dual effect. For Mode B, it eliminated the runaway cascade: expert 14, spawned at step 650, stabilized and was retained permanently. For Mode C, it *created* the conditions for replacement by exposing veterans at full fitness while temporarily shielding newborns. The parallel to infant industry tariffs in trade economics is precise: temporary, transparent, and declining protection.

### 3.3 Phase 3: The Round-Trip Domain Shift — Molecular Memory

The definitive experiment. Runs B2c and C2c executed two domain shifts: data_char → data_code at step 1500, then data_code → data_char at step 3000. The hypothesis: if old experts survive through the foreign domain, they should reactivate near-instantaneously when their original domain returns.

*Event counts by phase*

| Phase | B2c | C2c |
| --- | --- | --- |
| Steps 0–1500 (data_char) | 3 | 0 |
| Steps 1500–3000 (data_code) | 0 | 0 |
| **Steps 3000+ (data_char, return)** | 0 | 0 |

*Loss trajectory around domain shift 2*

| Step | B2c | C2c | Phase |
| --- | --- | --- | --- |
| 2800 | 1.150 | 1.157 | Pre-shift, data_code |
| 2950 | 1.270 | 1.126 | Last pre-shift steps |
| **3000** | *Domain shift 2 → data_char* | | |
| 3050 | 1.246 | 1.277 | +50 steps |
| 3100 | 1.204 | **1.140** | C2c < 1.15 |
| 3150 | **1.145** | 1.162 | B2c < 1.15 |
| 3200 | 1.089 | 1.094 | Both recovered |

| 3500 | 1.069 | 1.052 | Steady state |

*Recovery speed comparison*

| B2c (cost + grace) | C2c (exclusive + grace) |
|---|---|
| **9×** | **11.5×** |
| 150 steps vs. 1350 from scratch | 100 steps vs. 1150 from scratch |

> **Central result.** The system recovered to baseline performance in 100–150 steps compared to 1150–1350 from scratch — a 9–11× speedup. Zero births, zero replacements. The router simply redirected tokens back to surviving experts whose weights were intact from the original domain.

## 4. Discussion

### 4.1 Molecular Memory

The round-trip result reveals a property we term *molecular memory*, by analogy with chemical systems where reaction intermediates persist in a mixture even when not actively consumed, ready to participate when conditions change. In the FMA-orchestrated MoE:

**Experts are not destroyed** when their domain becomes inactive. The cost-penalized fitness formula reduces their score (low demand D → low fitness), but the replacement threshold is not crossed because the pool's fitness distribution remains relatively compact. **Expert weights are preserved** through the foreign domain phase. Unlike pruning or merging, the FMA keeps dormant experts' parameters intact. **Reactivation is near-instantaneous** because no new learning is required — the router discovers that old experts already produce low loss on the returning domain.

| Approach | Domain Memory | Multi-Domain Cost |
|---|---|---|
| Standard MoE (fixed routing) | None | Full retraining |
| Expert pruning | Destroyed | Regrowth required |
| Expert merging | Diluted | Quality loss on return |
| FMA-orchestrated MoE | Preserved | Near-zero adaptation |

### 4.2 The Grace Period as Infant Industry Protection

| Protection | Run | Outcome | Economic Analogue |
|---|---|---|---|
| None | B2 | Runaway loop (16) | Free trade crushes infants |
| Excessive (implicit) | C2 | Frozen market (0) | Permanent tariffs |
| **Calibrated (linear ramp)** | B2b/C2b | Balanced (6–7) | Temporary, declining tariff |

The economic parallel is precise: infant industry tariffs work when they are temporary, transparent, and declining — exactly the properties of the linear ramp. This mechanism, emergent from the FMA's market logic, has direct analogues in both international trade theory and the Evans-Polanyi barrier of chemical kinetics.

### 4.3 Mode C vs. Mode B

Mode C (exclusive_cost_grace) achieved zero expert turnover across the entire run including two domain shifts, adapting purely through gradient updates and router reweighting. This represents maximum expert pool stability. Mode B produced slightly more aggressive initial dynamics (three early replacements) but identical zero-churn behavior post-warmup. For production deployment, Mode C's conservatism may be preferable: it avoids any expert loss, maximizing the probability that all domain knowledge is retained.

## 5. Conclusions

| # | Finding | Evidence |
|---|---|---|
| 1 | **Cost-penalized fitness breaks the saturated-regime deadlock.** | B2: 16 replacements vs. A2: 6 (baseline) |
| 2 | **Linear grace periods are essential.** | B2b: 7 (tamed); C2b: 6 (unlocked) |
| 3 | **FMA-orchestrated MoE exhibits molecular memory.** | 9–11× faster recovery, zero churn on return |
| 4 | **Mode C is the most stable configuration.** | C2c: 0 replacements across entire run |
| 5 | **Estimated commercial value is substantial (see Appendix A).** | $39.1M/yr, 27.1 GWh reduction |

These results suggest that treating MoE expert management as a market optimization problem — where experts compete on cost-adjusted fitness rather than quality alone — produces emergent properties (memory, stability, efficient adaptation) that are difficult to engineer through conventional approaches. The FMA's 18-mechanism framework [1,2,10], originally developed for agent-based economic modeling, transfers to the MoE context with remarkable fidelity. Chemistry provides a particularly rich homology: the analogy between reaction intermediates and neural network experts predicts specific, testable behaviors that our experiments confirm quantitatively.

## Future Directions

**Scale validation:** Reproduce on larger MoE architectures (64+ experts, 12+ layers). There is reason to expect that the core mechanisms will transfer: the FMA's fitness evaluation operates on *normalized* per-expert statistics (quality ratios, demand fractions, relative FLOPs), and the replacement threshold (mean − 1σ) is a statistical criterion that is scale-invariant by construction. The grace period is parameterized in market steps, not expert counts. In economic terms, market dynamics do not depend on the number of firms — a market with 8 firms and one with 256 obey the same competitive logic. The only open question is whether the fitness *distributions* change shape at scale in ways that affect threshold crossings, which is precisely what scale experiments would determine.

**Multi-domain round-trip:** Extend to three or more domain cycles to test cumulative repertoire growth.
**Adaptive grace:** Temperature-dependent grace periods analogous to the Evans-Polanyi barrier.
**Inference-time FMA:** Apply market dynamics at inference with live cost/quality signals.

## Author Note

The author is not a machine-learning researcher. His background is in semiconductor physics simulation, followed by a decade in quantum-chemistry-based reaction kinetics modeling, and more recently in agent-based economic modeling. The FMA framework was developed over several years to address self-organization in complex economic systems [1,2,10], and was subsequently found to transfer to prebiotic chemistry, where its mechanisms map onto reaction kinetics. The present work extends that same framework to MoE expert management in transformers — a domain the author approaches as an outsider. Ongoing work applies the FMA to adaptive immune system modeling, where the parallels with expert lifecycle management are particularly direct: lymphocyte clonal selection maps onto expert competition, affinity maturation onto fitness-driven specialization, and immunological memory onto the molecular memory effect reported here. The contribution is therefore methodological rather than architectural: the paper proposes that the FMA's market-inspired population dynamics, already validated in economics, chemistry, and immunology, may offer a useful lens for managing expert lifecycles in large language models. The author hopes this cross-disciplinary perspective proves valuable to the MoE community.

## Acknowledgments

This research made extensive use of AI coding assistants (Anthropic Claude Opus and Sonnet) for implementing the nanoFMT codebase, running experiments on the university cluster, data analysis, and drafting this manuscript. The experimental design, theoretical framework, and interpretation of results are the author's own. The author gratefully acknowledges the University of Valladolid for computing resources.

# Appendix A. Preliminary Value Assessment for Commercial LLMs

This appendix presents a back-of-envelope estimate of the potential commercial impact. The figures below are intended as order-of-magnitude indicators to motivate further investigation, not as precise fore-casts. They rest on publicly available cost data and the experimentally observed speedup ratios from Sections 3–4.

Current large-scale MoE systems represent massive compute investments: DeepSeek-V3 (256 experts, 8 active/token, $5.6M training), Mixtral 8×22B (~$10–20M), and GPT-4 (rumored 16× expert clusters, >$100M). These systems serve billions of tokens daily across diverse domains. Domain adaptation represents a significant and recurring operational cost.

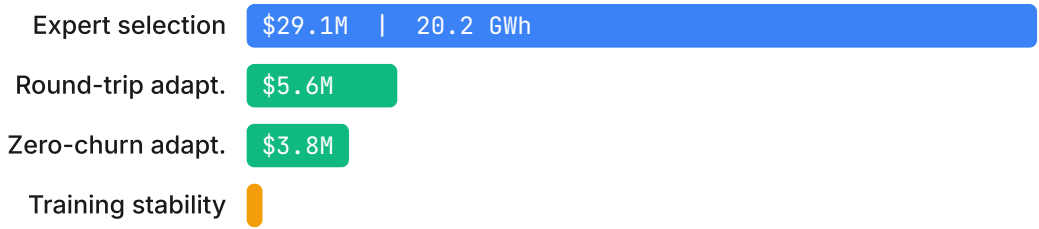

**Figure 3.** Estimated annual savings by FMA mechanism (moderate scenario, OpenAI-scale provider). Total: $39.1M, 27.1 GWh (~10,840 t $CO_2$).

**Total Annual Savings**

**$39.1M**

Moderate scenario

**Energy Reduction**

**27.1 GWh**

~10,840 t $CO_2$ avoided

The most distinctive FMA contribution is in *domain adaptation lifecycle management*. Current approaches to MoE efficiency (pruning, quantization, Expert Choice routing) optimize static performance. None address the dynamic cost of maintaining performance across shifting domains. The FMA's molecular memory creates a fundamentally different cost structure: a model trained on domains A, B, and C retains all three capabilities simultaneously. When domain A resurges, adaptation cost is near-zero. This compounding return on training investment has no equivalent in current practice.

**Caveats.** These estimates extrapolate from an 8-expert, 2-layer nanoFMT to production-scale systems with 256+ experts. The 9–11× re-adaptation speedup is measured; the dollar translation assumes linear scaling of the per-expert savings, which may be optimistic. Independent replication at scale (see Future Directions) is needed before these figures can be taken as reliable projections.